%% file: main.tex
\documentclass[10pt,twocolumn,letterpaper]{article}

\usepackage[pagenumbers]{cvpr} 

\input{preamble}
\definecolor{cvprblue}{rgb}{0.21,0.49,0.74}
\usepackage[pagebackref,breaklinks,colorlinks,allcolors=cvprblue]{hyperref}

\def\confName{CVPR}
\def\confYear{2027}

\usepackage[table]{xcolor}
\usepackage{hyperref}
\usepackage{url}
\usepackage{enumitem}
\usepackage{booktabs}
\usepackage{wrapfig}
\usepackage{tcolorbox}
\tcbuselibrary{skins, breakable}

\hypersetup{
  colorlinks=true,
  citecolor=blue!60!black,
  linkcolor=blue!60!black,
  urlcolor=blue!60!black
}

\newtcolorbox{promptbox}[2][]{
  enhanced,
  breakable,
  colback=white,
  colframe=black!70,
  colbacktitle=black!70,
  coltitle=white,
  title={#2},
  fonttitle=\bfseries,
  left=6pt, right=6pt, top=6pt, bottom=6pt,
  boxrule=0.6pt,
  arc=2mm,
  #1
}

\usepackage{pifont}
\usepackage{float}
\newcommand{\cmark}{\ding{51}}
\newcommand{\xmark}{\ding{55}}

\definecolor{LightBlue}{HTML}{B0E0E6}
\colorlet{LightBlue}{cyan!20}

\title{Video-as-Experience: Learning Computer Use via Audio-Visual Understanding}

\author{
Guangzhi Sun\\
University of Cambridge\\
Cambridge, UK\\
{\tt\small gs534@cam.ac.uk}
\and
Zhonghao Zhai, Yixuan Li, Chao Zhang\\
Tsinghua University\\
Beijing, China\\
{\tt\small {dizh23,liyixuan25,cz277}@tsinghua.edu.cn}
}

\begin{document}
\maketitle
\input{sec/0_abstract}    
\input{sec/1_intro}
\input{sec/2_related_work}
\input{sec/3_Method}
\input{sec/4_ExpSetup}
\input{sec/5_results}
\input{sec/6_conclusion}

\section{LLM Use Statement}

In this work, we used generative LLMs for generating descriptions for tutorial videos in our datasets, polishing paper writing and offering constructive feedbacks to the paper. We have not used generative AI tools for retrieval and discovery, research ideation or execution. Proving mathematical claims are is applicable to this work. Additionally, we used generative AI tools for critical feedback on the complete draft. We have reviewed all AI-assisted work. For example, we checked the polished texts word by word to ensure they still maintain our original intention and the key facts are correct. We take responsibility for the final content of this work, including text, claims or artifacts produced with the aid of generative AI.

{
    \small
    \bibliographystyle{ieeenat_fullname}
    \bibliography{main}
}

\input{sec/X_suppl}

\end{document}

%% file: sec/0_abstract.tex
\begin{abstract}
Multimodal large language models (LLMs) have enabled capable computer-use agents that perceive graphical user interfaces, reason over instructions, and execute actions. However, existing approaches to learning from video demonstrations typically convert videos into screenshot-based action trajectories, rather than directly leveraging their rich audio-visual information. We introduce VEX (Video as EXperience), a framework that uses audio-visual understanding to unify learning from external demonstrations and reflecting on an agent's own behavior. Following a watch-act-rewatch paradigm, the agent extracts procedural guidance from tutorial videos containing visual demonstrations and spoken explanations, executes tasks, and verifies its progress by rewatching its execution as video. We instantiate VEX with video-SALMONN-VEX, a 9B-scale audio-visual computer-use agent built on the video-SALMONN 2 architecture with a Qwen3.5-9B backbone. We further introduce Action-Grounded Guide Optimization (AGGO), a reinforcement learning method that optimizes video-derived guidance for downstream action prediction, and augment AgentNet, OSWorld, and CUA-Gym with over 24K real-world tutorial videos. Experiments on OSWorld and a complementary subset of CUA-Gym demonstrate consistent improvements, with video-SALMONN-VEX achieving a 22\% relative increase in task success over its Qwen3.5 backbone on OSWorld. Moreover, its generated guidance improves Gemini-3.5-flash by 10\% relatively on OSWorld. These results highlight the potential of audio-visual understanding as a unified interface for procedural learning and self-verification in computer-use agents.
\end{abstract}

%% file: sec/1_intro.tex
\section{Introduction}
\label{sec:intro}

Recent advances in multimodal large language models (LLMs) \citep{li2024llavaov,zhang2024video,lin2024vila,bai2025qwen25vltechnicalreport,damonlpsg2025videollama3,vsalmonn2,vsalmonno1,qwen3vl,vsalmonns} have enabled increasingly capable agents for computer use, where models perceive graphical user interfaces (GUIs), reason over task instructions, and execute actions in interactive environments \citep{koh2024visualwebarena,he2024webvoyager,cheng2024seeclick,rawles2024androidworld}. Most existing computer-use agents (CUAs), however, are primarily optimized to act from the information available in the current interaction, such as screenshots, textual instructions, and previously executed actions \citep{agashe2024agents,zhang2024ufo,gou2024uground,xu2024aguvis,qin2025uitars,gta1,fu2025mano,li2025dartgui,xue2026evocua,wang2025opencua}. In contrast, humans often acquire new procedural skills by observing demonstrations from experienced individuals. To this end, online tutorial videos provide an abundant record of such demonstrations, jointly capturing visual states, user operations, and spoken explanations. 
This has motivated a growing line of work that leverages instructional videos to improve computer-use agents, either at training time or as additional knowledge at inference time. Recent approaches \citep{song2026watch,lu2026videoagenttrek,xiong2026video2gui,liu2026learning} have demonstrated that internet-scale videos can indeed provide valuable information for computer use, typically by recovering executable interaction trajectories from observed demonstrations and using the resulting trajectories for training or in-context guidance.

Despite these advances, existing approaches largely use video as a source of trajectory extraction, rather than as a form of experience that an agent can directly understand and reason over. Raw videos are commonly passed through dedicated processing pipelines \citep{liu2026learning} that construct and filter trajectories before they can be consumed by the main agent. These representations are effective and provide explicit access to actions and intermediate states, but constructing them often requires dedicated data-collection and processing pipelines involving additional large models. Besides, converting videos into such representations often discards valuable audio and visual explanations. More fundamentally, demonstrations and an agent's own interactions are often represented differently, with external experience originating as video while self-generated experience is recorded as interleaved screenshots, actions, and textual traces. A single audiovisual model could instead use the same audio-visual perceptual interface to extract task-relevant knowledge from observed demonstrations and revisit its own execution, unifying learning from others and reflecting on its own behavior as reasoning over videos.

To achieve such agents, we introduce VEX (Video as EXperience), a framework for computer-use agents that has audio-visual understanding capability to directly learn procedural knowledge or perform self-verification on video recordings. At inference time, VEX follows a simple watch-act-rewatch loop, where the agent watches and extracts guidance directly from tutorial video recordings, and applies the guidance to its task execution. The agent also verifies its success after completion by rewatching its own execution represented as a video recording.
We instantiate VEX with video-SALMONN-VEX, a single 9B-scale agent built on video-SALMONN 2 \cite{vsalmonn2} architecture with Qwen3.5-9B LLM backbone that can understand audio-visual inputs while performing computer-use tasks. In addition, we introduce Action-Grounded Guide Optimization (AGGO) to train video-SALMONN-VEX to extract procedural knowledge from tutorial videos according to their utility for downstream action prediction. 

The VEX framework is evaluated on both OSWorld \citep{osworld} and a complementary subset of CUA-Gym \citep{cuagym} focusing on tasks that are not covered by OSWorld. Consistent improvements were obtained using VEX for video-SALMONN-VEX compared to its backbone Qwen3.5-9B and video-SALMONN 2 models, with a notable 22\% relative task success rate improvement on OSWorld. Moreover, using the VEX framework together with task guidance generated from video-SALMONN-VEX on Gemini-3.5-flash, a relative 10\% improvement was achieved on OSWorld. The main contributions of this paper are summarized as follows:

\begin{itemize}[leftmargin=*]
    \item We propose \textbf{VEX}, a framework that uses video as a shared interface for learning from demonstrations and self-evaluating agent execution. We introduce \textbf{video-SALMONN-VEX} as the single agent with audio-visual understanding capabilities that can learn from real-world video tutorials to perform the task and can verify its performance through video recordings of its execution.
    \vspace{0.1cm}
    \item We introduce \textbf{AGGO}, a reinforcement learning approach that trains the agent to extract procedural guidance from tutorial videos by rewarding guides according to their utility for predicting ground-truth action trajectories.
    \vspace{0.1cm}
    \item We augment AgentNet, OSWorld, and CUA-Gym with over {25K} real tutorial videos, and evaluate VEX on OSWorld and a complementary subset of CUA-Gym, where video-SALMONN-VEX achieves a \textbf{22\%} relative task success rate increase over the Qwen3.5-9B backbone.
\end{itemize}

%% file: sec/2_related_work.tex
\section{Related Work}
\label{sec:formatting}

\subsection{Computer-Use Agents}

Recent progress in multimodal LLMs has driven rapid advances in agents that interact with computers through graphical user interfaces. Early work studied web navigation and grounding from demonstrations and structured webpages, including WebShop, Mind2Web, and WebArena \citep{yao2022webshop,deng2023mind2web,zhou2024webarena}, while subsequent benchmarks such as VisualWebArena, WebVoyager, AndroidWorld, Windows Agent Arena, and OSWorld extended evaluation to visually grounded websites, mobile devices, and open-ended desktop environments \citep{koh2024visualwebarena,he2024webvoyager,rawles2024androidworld,osworld}. In parallel, GUI-specialized multimodal models and agent frameworks have improved visual grounding, action prediction, and long-horizon interaction, including CogAgent \citep{hong2024cogagent}, SeeClick \citep{cheng2024seeclick}, OS-ATLAS \citep{wu2025osatlas}, ShowUI \citep{lin2025showui}, Agent S \citep{agashe2024agents}, UFO \citep{zhang2024ufo}, Aguvis \citep{xu2024aguvis}, UI-TARS \citep{qin2025uitars}, GTA-1 \citep{gta1}, MANO \citep{fu2025mano}, DART-GUI \citep{li2025dartgui}, EvoCUA \citep{xue2026evocua}, and OpenCUA \citep{wang2025opencua}. These approaches primarily improve computer use through GUI-specific training data, interaction traces, grounding objectives, planning, or reasoning over the current observation and interaction history. VEX complements these by focusing on how procedural experience and an agent's past execution can be incorporated through a shared video interface.

\subsection{Learning from Video Experience}

Recent work moved beyond passive video understanding toward acquiring reusable knowledge and skills from prior video experience. SLVMBench \citep{slvmbench} studies this setting explicitly by evaluating whether models can extract procedural knowledge from tutorial videos retained in long-form video memory and later apply that knowledge to ongoing tasks, while video-SALMONN~S \citep{vsalmonns} provides a complementary architectural perspective by enabling selective retention and retrieval from multi-hour audiovisual streams. In computer use specifically, instructional videos have emerged as a scalable source of procedural experience: Watch \& Learn \citep{song2026watch} recovers executable interaction trajectories from online demonstrations for training and in-context guidance, while related approaches convert, segment, retrieve, or filter tutorial videos into action trajectories or procedural guides for downstream GUI interaction \citep{lu2026videoagenttrek,xiong2026video2gui,liu2026learning,zhang2026tongui}. These studies demonstrate that videos contain valuable task knowledge, but generally transform demonstrations into an intermediate representation before the acting agent consumes them. VEX instead treats video itself as the common representation of experience: the same audio-visual agent can watch external demonstrations to acquire task-relevant procedural knowledge and rewatch recordings of its own execution for verification. 

%% file: sec/3_Method.tex
\section{Methodology}
\label{sec:vex}

\begin{figure*}[t]
    \centering
    \includegraphics[width=\linewidth]{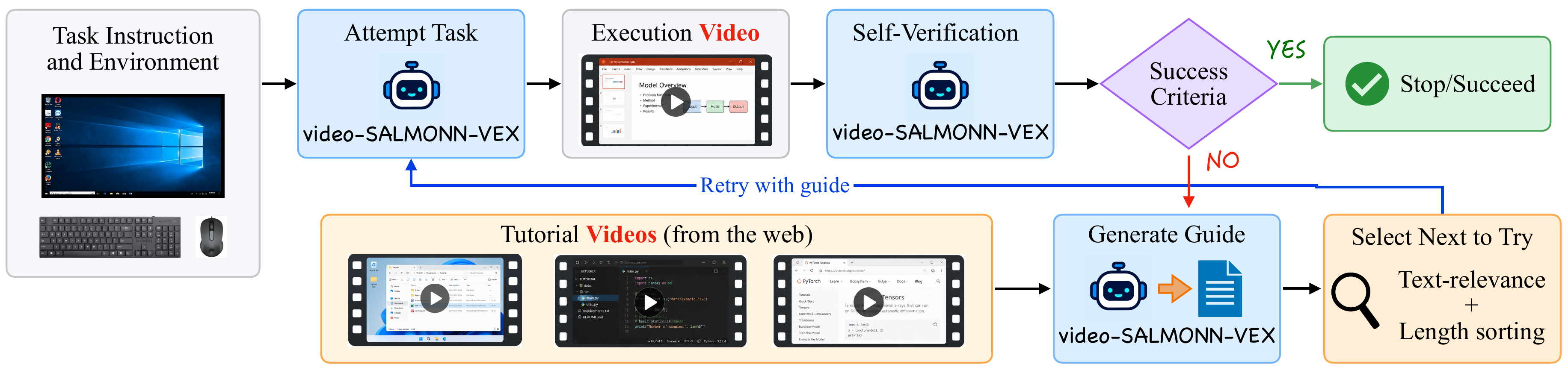}
    \caption{Illustration of the \textbf{Video-as-EXperience (VEX) framework}. We highlight interfaces that employ video as the representation. The VEX framework starts with task execution without video first, followed by self-verification based on a video recording of its execution. If the execution failed to meet criteria, the agent will fetch the next most informative video tutorial, extract guidance and re-perform the task. The loop is repeated until success or the maximum trial is reached.}
    \label{fig:vex}
\end{figure*}

\begin{figure*}[t]
    \centering
    \includegraphics[width=\linewidth]{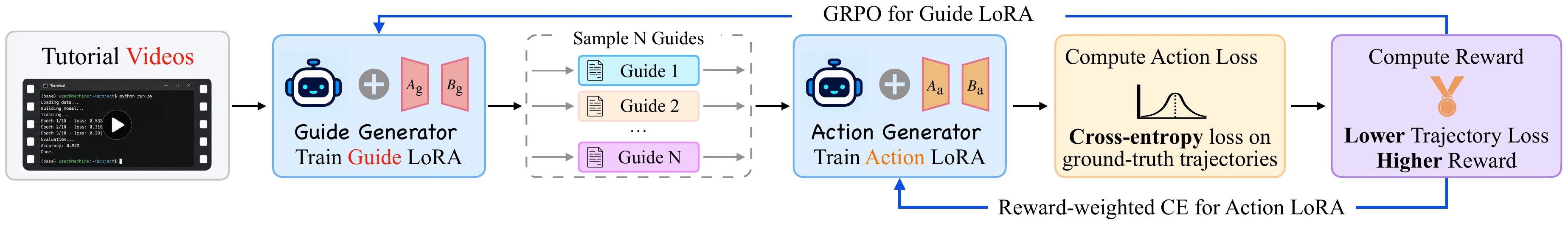}
    \caption{Illustration of the \textbf{Action-Grounded Guide Optimization (AGGO)} training pipeline for video-SALMONN-VEX. The guide generation policy as a LoRA adapter is trained with GRPO, and the action generation policy as a separate LoRA adapter is trained with weighted CE loss, as described in Section \ref{sec:aggo}.}
    \vspace{-0.3cm}
    \label{fig:aggo}
\end{figure*}

\subsection{VEX Framework Overview}

Fig. \ref{fig:vex} shows an overview of the VEX framework. 
The core idea is to use a single agent with audio-visual understanding capabilities, such as video-SALMONN-VEX, to learn from video, perform actions, and verify its action trajectory represented as a video. A loop is formed when the verification result does not meet the pass criterion, enabling a second trial with more knowledge from video.

Specifically, starting with the task instruction and environment, the agent performs the task without any further inputs for the first time, and the trajectory is recorded as videos with action descriptions converted to audio via text-to-speech (TTS) synthesis. After re-watching its own actions, a self-verification score is provided by the agent itself (see Section \ref{sec:selfverify} for detailed implementations). Then, a threshold check is performed, and when the execution is below the threshold, the agent will do the task again with the help of external video tutorials. 

\begin{figure*}[t]
    \centering
    \includegraphics[width=\linewidth]{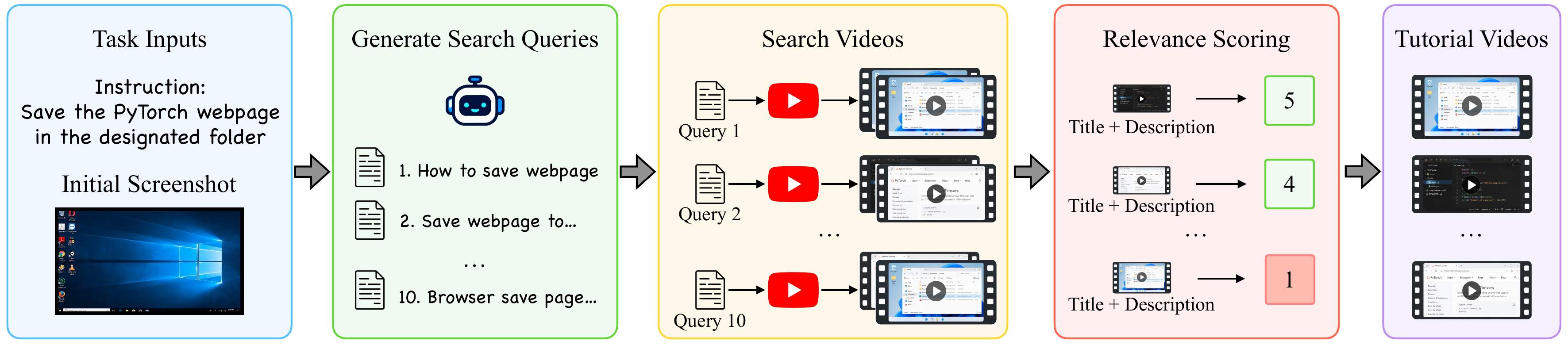}
    \caption{Illustration of the search pipeline to get the candidate pool of tutorial videos. The agent used in this pipeline is the same agent used in the VEX framework. The scoring is based on title and description, with a relatively coarse 1 to 5 score scale.}
    \label{fig:videosearch}
\end{figure*}

To learn from video tutorials, a pool of candidate videos is first obtained following the pipeline shown in Fig. \ref{fig:videosearch}, which is constructed with the help of the same agent itself. Then, guidance texts are extracted from each video recording directly by providing the task description and the initial screenshot to the agent. The agent is prompted to extract as much information as possible, while being honest to respond \textit{no information found} when the video does not contain useful procedural knowledge. Prompts are provided in Section \ref{sec:prompt} in the supplementary materials. Then, when the first trial fails, the video with the longest extracted guidance is selected, and the guidance extracted by the agent is used as the context for the agent to perform the task. The agent then goes through the act-and-rewatch procedure again, and once it fails, the second-longest video is used to provide tutorial knowledge. If the set maximum number of trials is reached and there is still no trajectory above the threshold, the one with the highest self-verification score is used as the final execution of the task.

\subsection{video-SALMONN-VEX and AGGO Training}
\label{sec:aggo}

We introduce video-SALMONN-VEX, which is an audio-visual LLM capable of understanding both audio and visual inputs in videos. The auditory capability is particularly important since most instructional information is conveyed via speech in demonstration videos. The model structure inherits from video-SALMONN 2 \citep{vsalmonn2}, and we subsequently fine-tune its LLM backbone with low-rank adapters (LoRA) \citep{hu2022lora} via the AGGO algorithm introduced as follows.

The process of AGGO is illustrated in Fig. \ref{fig:aggo}, which leverages data containing ground-truth trajectories to train the guidance from videos via RL. Specifically, we train the model to extract task-relevant textual guidance from tutorial videos leveraging the audio-visual understanding capability of the agent, and use the generated guidance to improve downstream action prediction. Each training example consists of a task instruction $q$, an instructional video $V$, and an oracle computer-use trajectory $\tau^{*}=\{(o_1,a_1^{*}),\ldots,(o_T,a_T^{*})\}$, where $o_t$ and $a_t^{*}$ denote the observation and groundtruth action at step $t$, respectively. Rather than supervising the generated guidance with a reference, we directly evaluate its usefulness based on how well it supports prediction of the ground-truth trajectory.

Two separate LoRA adapters are used, denoted as $\phi_{\mathrm{g}}$ for the guidance generation policy and $\phi_{\mathrm{a}}$ for the action prediction policy, since they require different capabilities. The action policy is first initialized from supervised fine-tuning (SFT) using ground-truth trajectories without guidance, and the guidance policy is initialized from SFT using descriptions generated by a strong model (e.g. Gemini-3.5-flash). The two adapters have disjoint trainable parameters while sharing the same frozen backbone. For each task-video pair, we sample $N$ candidate responses of step-by-step guidance from the guidance generation policy,
\begin{equation} 
g_i \sim \pi_{\mathrm{g},\phi_{\mathrm{g}}}(\cdot \mid q,V), \qquad i=1,\ldots,N. 
\end{equation}
Each candidate guidance is then evaluated using the action policy on the \emph{entire} ground-truth trajectory. Specifically, we teacher-force the full trajectory and compute its sequence-level log-likelihood conditioned on $g_i$ as shown in Eqn. (\ref{eq:trajectory_reward})
\begin{equation} 
\label{eq:trajectory_reward} 
r_i = \frac{1}{T} \sum_{t=1}^{T} \log \pi_{\mathrm{a},\phi_{\mathrm{a}}} \left( a_t^{*} \mid q,o_{\leq t},a_{<t}^{*},g_i \right). 
\end{equation}
Thus, each generated guidance receives a single scalar reward determined by the likelihood of the complete ground-truth trajectory. The reward measures the overall utility for explaining the demonstrated task execution. Guidance that makes the full ground-truth trajectory more likely under the action policy receives a higher reward. When computing this reward, the action prediction policy is used only as a scorer, and gradients are not propagated through it to the guidance generation policy.

We optimize the guidance LoRA using Group Relative Policy Optimization (GRPO), with rewards normalized using a leave-one-out (RLOO) \citep{ahmadian2024back} baseline within the $N$ guidance samples generated for the same task--video pair. The advantage of candidate $g_i$ is
\begin{equation} 
\label{eq:rloo} 
\hat{A}_i = r_i - \frac{1}{N-1} \sum_{\substack{j=1, j\neq i}}^{N} r_j. 
\end{equation}
The resulting sequence-level advantage is assigned to the tokens of the corresponding text guide and optimized using the standard clipped GRPO \citep{shao2024deepseekmath} objective. The RLOO advantage computation provides a direct relative comparison between alternative guidance while reducing sensitivity to differences in the task difficulty.

Simultaneously, we train the action policy LoRA to better utilize the sampled guidance. Instead of treating all generated guidance equally, we use their trajectory-level rewards to assign a normalized importance weight below.
\begin{equation} 
\label{eq:guidance_weight} 
w(g_i) = \text{Softmax}(g_i / T_w), \end{equation} 
where $T_w$ is a temperature controlling the concentration of the weighting distribution. The action policy is trained with weighted cross-entropy (CE) as shown in Eqn. (\ref{eq:action_ce}).
\begin{equation} 
\label{eq:action_ce} 
\mathcal{L} = -\sum_{i=1}^{N} w(g_i) \left[ \frac{1}{T} \sum_{t=1}^{T} \log \pi_{\mathrm{a},\phi_{\mathrm{a}}} \left( a_t^{*} \mid q,o_{\leq t},a_{<t}^{*},g_i \right) \right]. \end{equation}
This weighting couples the two learning objectives. Consequently, the guidance LoRA is encouraged to generate video-derived information that better supports the complete oracle trajectory, while the action LoRA is preferentially trained on guidances that are already more useful for downstream action prediction. Moreover, instead of using the entire dataset, which has been used in SFT for the action policy, we first perform per-step action prediction to select the top 30\% of trajectories that are more error-prone to be used as the training set for RL.

\begin{figure*}[t]
    \centering
    \includegraphics[width=0.9\linewidth]{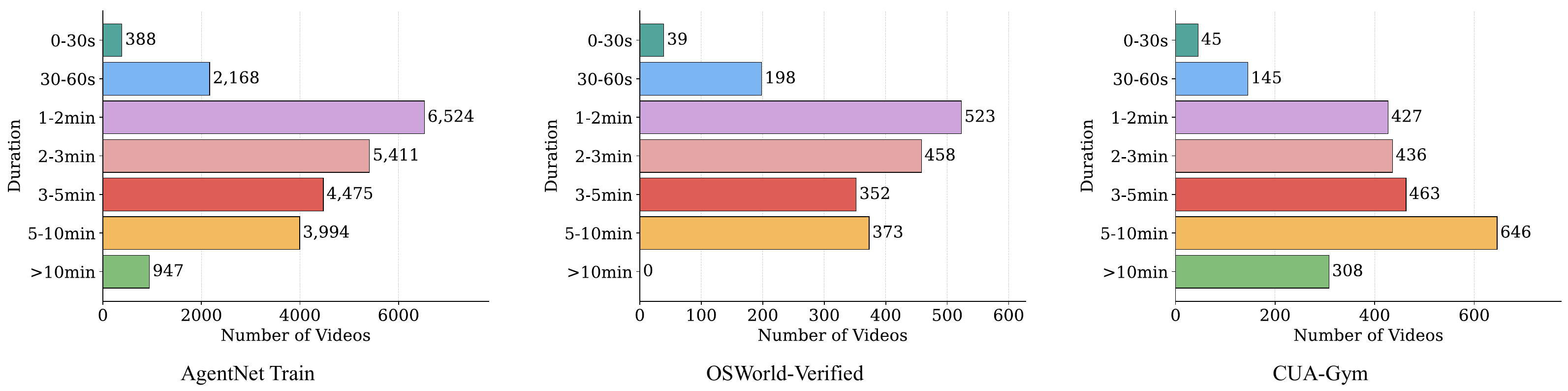}
    \vspace{-0.2cm}
    \caption{Distribution of durations for tutorial videos in the AgentNet training set, the OSWorld-Verified and the CUA-Gym subset.}
    \vspace{-0.3cm}
    \label{fig:train_dist}
\end{figure*}

\subsection{Video-based Self-verification}
\label{sec:selfverify}

To perform self-verification using video as the only interface, we synthesise videos from trajectories by joining screenshots together, where each screenshot occupies a duration determined by the duration of the synthesized speech of the action description\footnote{The VEX framework prompts the agent to generate a description of the action followed by the code.}. This way, the pipeline simulates the situation where the agent inspects the demonstration video it recorded to judge its performance.

Given each execution video, we prompt the model to give an integer score of 0 to 9, with detailed grades specified in the supplementary material. Then, following \cite{llmverify}, we compute the expected score by averaging the token probabilities with their values as shown in Eqn. (\ref{eq:verifier}).
\begin{equation}
    s_\tau = \sum_{c=0}^9 cP(c|V_\tau, q; \theta)
    \label{eq:verifier}
\end{equation}
where $V_\tau$ denotes the video of the trajectory $\tau$, and $\theta$ denotes the backbone video understanding model without the two LoRA-based policies. This scoring mechanism is more robust than simply choosing the score with the highest probability, especially when the boundaries between adjacent grades are inevitably ambiguous and the model assigns high probabilities to more than one score.

%% file: sec/4_ExpSetup.tex
\section{Experimental Setup}
\label{sec:exp_setup}

\subsection{Video-augmented Datasets}

In addition to the video-SALMONN 2 training set, the video-SALMONN-VEX model uses the AgentNet dataset \citep{wang2025opencua} as the only CUA dataset for fine-tuning. As part of our contribution, we augment AgentNet with 20k tutorial videos covering 7k tasks. These videos are selected using the same video selection pipeline illustrated in Fig.~\ref{fig:videosearch}. For each video-task pair, we collect reference guidance text generated by Gemini-3.5-flash, which can be used for cold-start fine-tuning of the guidance generation policy LoRA adapter. The duration distribution of these videos in AgentNet is shown in Fig~\ref{fig:train_dist}.

We primarily evaluate agents on OSWorld-Verified \citep{osworld} using all tasks. To assess the generalization of video-SALMONN-VEX, we additionally evaluate on a subset of 220 tasks from CUA-Gym \citep{cuagym} that complements the OSWorld tasks. There are 49 PDF tasks from CUA-Gym incorporated since this requires software not covered by OSWorld-Verified. The distribution of each task category in the CUA-Gym subset is given in Table \ref{tab:cuagym}. Following the video collection pipeline, we obtain 2k tutorial videos for OSWorld and 3k for CUA-Gym from YouTube that are under 10 mins. The distribution of durations for videos after the relevance filtering is also provided in Fig. \ref{fig:train_dist}.


\begin{table}[t]
    \centering
    \footnotesize
     \caption{Number of tasks in each task category for the CUA-Gym test data used in our experiments. CUA-Gym contains only 2 OS-related tasks, so we omit them in this plot. PDF is not in OSWorld.}
    \begin{tabular}{p{5cm}c}
    \toprule
    Task Type     & Task Count \\
    \midrule
     Office    & 94 \\
     Daily &  20 \\
     Professional & 39 \\
     Work &  16 \\
     PDF & 49 \\
     \bottomrule
    \end{tabular}
    \vspace{-0.2cm}
    \label{tab:cuagym}
\end{table}

\begin{table*}[t]
    \centering
    \caption{Success rate on OSWorld-Verified using video-SALMONN-VEX compared to its backbone models, including Qwen3-VL, Qwen3.5, video-SALMONN 2 and video-SALMONN 2 SFT. We use results across maximum 3 attempts for video-SALMONN 2 SFT @ 3 trials using the same verification as VEX, and use the VEX framework for video-SALMONN-VEX with a maximum of 3 trials.}
    \resizebox{0.9\textwidth}{!}{%
    \begin{tabular}{p{8cm}ccccc|c}
    \toprule
    Models & OS & Office & Daily & Professional & Workflow & Overall\\
    \midrule
    Qwen3-VL (8B) \citep{qwen3vl}                         & 35.15 & 28.57 & 46.24 & 65.31 & 20.82 & 36.52 \\
    Qwen3.5 (9B) \citep{qwen3vl}                         & 62.50   & 36.72 &  46.80 &   67.35 &  22.77 &  41.15 \\
    Qwen3.5 (9B) + Trajectory \citep{liu2026learning} & 64.90 & 35.21 & 54.62 & 66.45 & 29.47 & 43.98 \\
    video-SALMONN 2 (9B) \citep{vsalmonn2}                & 54.17  & 36.72 &  51.24 &  73.47 &  24.05 &  42.70 \\
    video-SALMONN 2 SFT (9B)            &  54.17  & 38.45 & 56.95  & 75.51 & 19.45 & 43.46 \\
    video-SALMONN 2 SFT (9B) @ 3 trials  & 63.19 & 37.14 & 61.68 & 72.45 & 25.20  & 45.65 \\
    \midrule
    \rowcolor{LightBlue} video-SALMONN-VEX (9B) (ours) & 50.00    & {47.86} & {62.67} & {81.63} & {25.93} & \textbf{49.99}\\
    \bottomrule
    \end{tabular}}
    \label{tab:osworldmain}
\end{table*}

\begin{table*}[t]
    \centering
    \caption{Success rate on CUA-Gym using video-SALMONN-VEX compared to its backbone models, including Qwen3-VL, Qwen3.5, video-SALMONN 2 and video-SALMONN 2 SFT. We use results across maximum 3 attempts for video-SALMONN 2 SFT @ 3 trials using the same verification as VEX, and use VEX framework for video-SALMONN-VEX with maximum 3 trials.}
    \resizebox{0.9\textwidth}{!}{%
    \begin{tabular}{p{8cm}ccccc|c}
    \toprule
    Models & Office & Daily & Professional & Workflow & PDF & Overall\\
    \midrule
    Qwen3-VL (8B) \citep{qwen3vl}    & 8.94 & 22.00 & 25.51 & 14.81 & 2.04& 11.87 \\
    Qwen3.5 (9B) \citep{qwen3vl}    &12.32 & 36.00  &   28.46  & 15.54 &  12.76 & 18.33 \\
    Qwen3.5 (9B) + Trajectory \citep{liu2026learning} & 14.37 & 37.91 & 31.72 & 15.47 & 19.07 & 21.25 \\
    video-SALMONN 2 (9B) \citep{vsalmonn2}    & 13.16  & 20.00     & 22.44 &  24.17 & 15.41 & 17.23  \\
    video-SALMONN 2 SFT  (9B)   & 15.67 &  23.50 &  25.64 &  22.31 & 17.45 & 19.34 \\
    video-SALMONN 2 SFT (9B) @ 3 trials & 17.25 & 30.00  &   31.92 & 15.42 & 13.97 &  20.29  \\
    \midrule
    \rowcolor{LightBlue} video-SALMONN-VEX (9B) (ours) &  {21.96} & 41.00  & {36.03} & 17.50  & {22.61} & \textbf{26.26}  \\
    \bottomrule
    \end{tabular}}
    \label{tab:cuagymmain}
\end{table*}

\subsection{Model, Training and Inference Configurations}

We first train the video-SALMONN 2 architecture initialized with a pre-trained Qwen3.5-9B and the Whisper encoder, following the training pipeline described in \cite{vsalmonn2}. We denote this model as \textit{video-SALMONN 2} throughout the paper. Then, the LoRA adapter for the action generation policy, which is applied only to the LLM backbone, is first fine-tuned on the full AgentNet dataset for 2 epochs starting from video-SALMONN 2. We denote the resulting model as \textit{video-SALMONN 2 SFT}.
Separately, the LoRA adapter for the guidance generation policy is also fine-tuned with LoRA on the LLM backbone using the reference guidance texts generated by Gemini-3.5-flash, with the same video-SALMONN 2 backbone. Both adapters use a rank of 64 and an alpha value of 128. Fine-tuning the two adapters takes 12 hours in total on 32 H100 GPUs.

Next, we generate an action at each step of every training sample, conditioned on the oracle action history, and compute the proportion of steps with incorrect predictions. We select the 30\% of training samples with the highest error rates for AGGO training. For each video-task pair, we generate 9 guidance samples and set $T_w=0.2$. AGGO training runs for 2 epochs on this selected subset and takes 48 hours on 32 H100 GPUs. We take the last model checkpoint at the end of epoch 2 across our experiments. Training and inference code is provided at \url{https://github.com/TomZ1412/VideoCUA/tree/main}

The tutorial videos are processed at 1FPS with a maximum number of 768 frames at a resolution of 360p. The trajectory videos are made using Kokoro TTS and \texttt{ffmpeg} to glue screenshots and audios together. Unless otherwise specified, VEX uses a maximum of 3 trials and a self-verification threshold of 7.0 during inference. We examine both hyperparameters in the ablation study. If the available tutorial videos are used up before the trial limit is reached and the self-verification score remains below the threshold, the agent uses its remaining trials without tutorial videos. Each trial starts by re-configuring the machine to the beginning status. Execution stops when either the trial limit is reached or the self-verification score meets the threshold. If that happens, the final state of the trial with the highest verification score is used for scoring.

We also apply VEX to Gemini-3.5-flash, a larger and stronger proprietary model, to evaluate the generalization of VEX to stronger agents. Gemini-3.5-flash receives guidance generated either by itself or by the trained video-SALMONN-VEX model, allowing us to evaluate whether AGGO training improves the extraction of actionable skills from videos. For task execution with both video-SALMONN-VEX and Gemini-3.5-flash, we adapt the official OSWorld-Verified harness for Qwen models. When failure happens, we restart the task from the initial setup. For both models, we set the maximum steps within a single trial to be 50, with a history number of steps to be 5. Screenshots are taken and processed at 1080p resolution.

\subsection{Baseline Setups}

To evaluate the effectiveness of VEX, we compare it with two single-agent baselines: video-SALMONN 2 and its underlying Qwen3.5-9B model. This comparison ensures the same LLM backbone and the same supervised training data, making the VEX pipeline and the AGGO training the only variables to compare. We additionally include a multi-trial baseline in which each task is attempted at maximum three times and are verified using the same verification. This is the comparison to show the effectiveness of learning guidance from tutorial videos under the same budget. Finally, following \cite{liu2026learning}, we use QWen3.5 to convert tutorial videos into step-by-step textual trajectories and include these trajectories in the context of Qwen3.5 to perform tasks. This provides a stronger baseline that uses experience extracted from online videos. Ablation studies were performed in Section \ref{sec:ablation} where: (i). the isolated effect of AGGO training, and (ii). same multiple trial budgets without video tutorials were examined with their respective baselines.

%% file: sec/5_results.tex
\section{Results}
\label{sec:results}

\subsection{video-SALMONN-VEX Main Results}

\begin{table}[t]
    \centering
    \caption{Influence of audio modality in video-SALMONN-VEX. Guidance text is always provided, but is either extracted from silent videos or normal videos as indicated.}
    \small
    \begin{tabular}{lccc}
    \toprule
    Model     & Audio & OSWorld & OpenCUA \\
    \midrule
    Qwen3.5 & \xmark & 43.98 & 21.25 \\
    video-SALMONN-VEX & \xmark & 46.42 & 24.16 \\
    video-SALMONN-VEX & \cmark & \textbf{49.99} & \textbf{26.26} \\
    \bottomrule
    \end{tabular}
    \label{tab:audiovisual}
\end{table}

\begin{table}[t]
    \centering
    \footnotesize
    \caption{Ablation study on key design of video-SALMONN-VEX (denoted as vS-VEX) on OSWorld. When the video tutorial is not used but self-verification is turned on, we randomly sample 3 trajectories and perform verification to select the best-performing trajectory. When using video tutorial for video-SALMONN 2 SFT (denoted as vS-2 SFT), we directly use the tutorial video as the context without extracting guidance.}
    \begin{tabular}{lcccc}
    \toprule
    Model     & Video Tutorial & Self-Verification & Success Rate \\
    \midrule
    vS-VEX & \cmark & \cmark & 49.99 \\
    vS-VEX & \cmark & \xmark & 45.58  \\
    vS-VEX & \xmark & \cmark & 45.63 \\
    vS-VEX & \xmark & \xmark & 43.46 \\
    vS-2 SFT & \cmark & \xmark & 38.71 \\
    vS-2 SFT & \xmark & \xmark & 43.46 \\
    \bottomrule
    \end{tabular}
    \label{tab:ablation}
\end{table}

\begin{table*}[t]
    \centering
    \small
    \caption{Comparison between video and trajectory (screenshot plus text description) as the representation of the trial for self-verification. Pearson Correlation Coefficient (PCC) is between the predicted verification score and the per-task actual score, and AUROC is computed by converting the per-task actual score into binary with a threshold of 0.5.}
    \begin{tabular}{p{6cm}p{5cm}ccc}
    \toprule
    Representation   & Model for Verification  & PCC & AUROC & Success Rate \\
    \midrule
    Screenshot + Text Description & Qwen3.5 & 0.4077 & 0.7663 & 48.59 \\
    Video & Gemini-3.5-flash & \textbf{0.5416} & {0.7890} & 48.85 \\
    Video & video-SALMONN-VEX & 0.5218 & \textbf{0.8210} & \textbf{49.99}  \\
    \bottomrule
    \end{tabular}
    \label{tab:representation}
\end{table*}

\begin{table*}[t]
    \centering
    \small
    \caption{Success rate on OSWorld and GUA-Gym using \textbf{Gemini-3.5-flash} as the action prediction model with the VEX framework. No tutorial means we perform multiple trials with the VEX framework without tutorial videos.}
    \begin{tabular}{p{6cm}p{5cm}cccc}
    \toprule
    Guidance Generation Model & Method & OSWorld & CUA-Gym & Overall\\
    \midrule
    Gemini-3.5-flash & Standard & 61.69 & 47.13 & 56.07 \\
    Gemini-3.5-flash & 3 trials & 63.60 & 45.07 & 56.45 \\
    Gemini-3.5-flash & VEX (no tutorial) & 67.72 & 50.12 & 60.93 \\
    Gemini-3.5-flash  & VEX & 67.38 & \textbf{52.70} & 61.71 \\
    \midrule
    \rowcolor{LightBlue} video-SALMONN-VEX & VEX  & \textbf{68.67} & 52.34 & \textbf{62.37} \\
    \bottomrule
    \end{tabular}
    \label{tab:osworldgemini}
\end{table*}

\begin{figure*}[t]
    \centering
    \includegraphics[width=\linewidth]{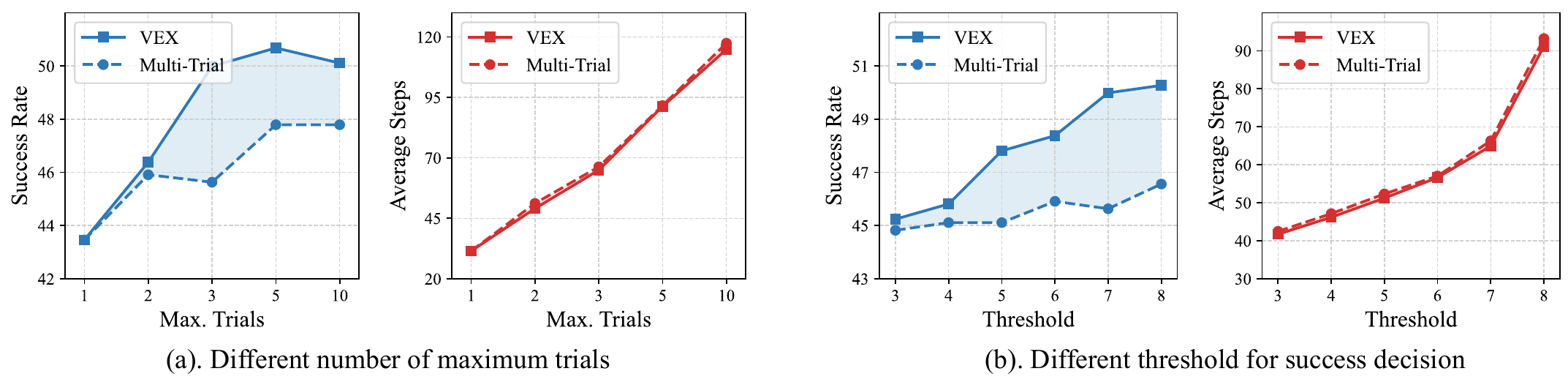}
    \caption{Change in success rate and average steps per task for video-SALMONN-VEX and its multi-trial counterpart using video-SALMONN 2 SFT, under (a) different maximum number of trial settings and (b) different threshold of verification score for task success. The multi-trial counterpart does not use tutorial videos, only keep trying until the threshold is met or the maximum trials is reached.}
    \label{fig:vex_step_plot}
\end{figure*}

Tables \ref{tab:osworldmain} and \ref{tab:cuagymmain} summarize the main results on OSWorld-Verified and CUA-Gym, respectively. The categorization of tasks is based on the OSWorld-Verified official split \citep{osworld}. video-SALMONN-VEX achieves the highest overall success rate among the evaluated methods on both benchmarks, reaching 49.99\% on OSWorld and 26.26\% on CUA-Gym. Compared with video-SALMONN 2 SFT, these results represent relative success rate gains of 15\% and 36\%, respectively. VEX also outperforms the underlying Qwen3.5 backbone by 22\% relatively on OSWorld-Verified and 43\% on CUA-Gym. The improvements across both benchmarks support the effectiveness of VEX for computer-use tasks, including the complementary tasks evaluated in CUA-Gym.

\textbf{The benefits of video demonstrations} are supported by comparing both Qwen3.5 + Trajectory against Qwen3.5, and comparing video-SALMONN-VEX with other baselines that do not use video tutorials. Similar levels of improvement as reported in \cite{liu2026learning} are observed on Qwen3.5, and much larger improvements are obtained using video-SALMONN-VEX, compared to the counterpart with the same trial budget but not using tutorial videos.


\textbf{Importance of audio-visual understanding}. Audio modality is crucial to understanding and learning from tutorial videos, which is evidenced in Table \ref{tab:audiovisual}. While all agents use guidance from videos, guidance extracted with audio perception achieves 6\% and 5\% absolute performance improvements respectively, demonstrating the core role of audio in explaining the procedures in the tutorial video.


\subsection{Ablation Studies}
\label{sec:ablation}

\textbf{Different functionality of VEX components} are shown in Table \ref{tab:ablation}, including the video tutorials and the self-verification on OSWorld. Both removing self-verification and removing video tutorials degrade the performance, and disabling both components together results in an absolute 6.53\% degradation. Both components therefore contribute to performance, with their combination outperforming either component alone. 
In addition, directly providing tutorial videos as multimodal context to video-SALMONN 2 SFT decreases success by 4.7\%. This suggests that access to videos alone is insufficient for a 9B-scale model, and that extracting actionable guidance is important for making effective use of tutorial content. Self-verification provides an additional improvement through trajectory selection.

\textbf{Video guidance improves the effectiveness and efficiency of repeated trials.} Fig. \ref{fig:vex_step_plot} compares VEX with a multi-trial baseline that uses the same self-verification loop without tutorial videos, varying the maximum number of trials and the verification threshold. Increasing the max trial from 1 to 3 substantially improves success rate, achieving substantially better performance compared to merely trying multiple times with the same verification process. The gain peaks at 5 trials, and then plataeus as the tutorial videos run out where the rest of the trials are bootstrapped by repeated trials without additional information. Moreover, VEX framework consistently achieves a lower average number of steps per task compared to simple multi-trial baseline. Specifically, at a maximum of 3 trials, VEX framework requires 64 steps on average, whereas the multi-trial baseline requires 67 steps. Meanwhile, the number of trials needed to achieve the same performance is much smaller with VEX. Across all tested thresholds, VEX achieves higher success rates with fewer steps.

\textbf{Video-based self-verification yields higher task success than screenshot-and-text verification.} The comparison between video-based verification in VEX, and trajectory-based (screenshot + text) verification, is shown in Table \ref{tab:representation}. We deliberately choose Qwen3.5 instead of video-SALMONN 2 for screenshot + text verification, since it generally performs better in understanding screenshots compared to a video-specific model, and any improvements compared to Qwen3.5, will be more salient when compared against video-SALMONN 2. Overall, video-based verification achieves better success rate with an absolute 1.40\% improvement than trajectory-based counterparts. 
The video-based configuration also achieves higher PCC with the actual task scores (0.5218 vs.\ 0.4077) and AUROC (0.8210 vs.\ 0.7663). Using Gemini-3.5-flash as the video verifier achieves further improvements on PCC but lower AUROC, and as a results yielding a slightly lower success rate compared to video-SALMONN-VEX.

\begin{table}[t]
    \centering
    \small
    \caption{Effect of dual-LoRA AGGO compared with single-LoRA AGGO and guidance SFT using Gemini-generated descriptions.}
    \label{tab:aggo}
    \setlength{\tabcolsep}{4pt}
    \begin{tabular}{@{}lcc@{}}
        \toprule
        Training Setting & Success Rate & Difference \\
        \midrule
        AGGO & 49.99\% & -- \\
        AGGO (single LoRA) & 43.72\% & $-6.27$\% \\
        Guidance SFT & 44.94\% & $-5.05$\% \\
        \bottomrule
    \end{tabular}
\end{table}

\textbf{Dual-LoRA AGGO outperforms both single-LoRA training and guidance SFT.}
Table~\ref{tab:aggo} shows that AGGO with both LoRA adapters achieves the highest success rate across the table, exceeding guidance SFT with Gemini-generated descriptions by 5.05\%. This improvement indicates that AGGO provides benefits beyond supervised imitation of reference guidance. Replacing the dual-LoRA design with a single adapter reduces success by 6.27\% absolute value, showing that the learning from tutorial videos and acting in the environment requires different set of skills to be represented in different adapters.

\subsection{Transferability to Gemini-3.5-flash}

Table~\ref{tab:osworldgemini} shows that VEX with Gemini-generated guidance improves overall success by an absolute 5.7\% using Gemini as the action agent, exceeding both the three-trial baseline and VEX without tutorials. Replacing this guidance with descriptions generated by video-SALMONN-VEX further increases the overall success rate by 0.7\%, demonstrating that its learned guidance can even benefit a stronger proprietary agent. Overall, a 10\% relative improvement in success rate on both test sets is achieved, demonstrating the effectiveness of VEX framework, and the knowledge gain by learning from tutorial videos.
Example guidance generated by Gemini-3.5-flash and video-SALMONN-VEX can be found in Section \ref{sec:guidance} in the supplementary materials.

%% file: sec/6_conclusion.tex
\section{Conclusion}

We introduced VEX, a watch, act and rewatch framework that uses video as a unified representation for learning from demonstrations and verifying an agent's own behavior. We instantiated it with the 8B-scale video-SALMONN-VEX, proposed AGGO to optimize guidance for downstream action prediction, and augmented three datasets with over 24K tutorial videos. Experiments demonstrate improved computer-use performance and transfer to Gemini-3.5-flash, highlighting the potential of video experience for procedural learning and self-verification.

%% file: sec/X_suppl.tex
\clearpage
\setcounter{page}{1}
\maketitlesupplementary

\section{Prompt Templates}
\label{sec:prompt}

\subsection{Video Guidance Extraction and Apply}

\begin{promptbox}[label={real_scripts}]{Video Guidance Extraction Prompt}
$<$tutorial video input$>$\\

You are given the above video tutorial teaching you how to perform some computer use tasks.\\

You will also be given a task description.\\

Use ONLY the knowledge from the video, describe how to perform the task step by step. \\
The video may not contain the entire task. It may only contain a part of the task. Just describe the part of the task that is shown in the video.
\\

For example, for a long task where part of it requires you to login with username and password that is included in the video, you should output the following:\\
When you want to log in, you should first click the login button on the left menu bar. Then, this will direct you to the login page. You need to enter your username and password. You should first click on the textbox with a word "Username" and then enter your username. Then, click on the textbox with a word "Password" and enter your password.
\\

\#\#\# Rules\\
1. Do not use your own knowledge. Just describe what you can learn from the video.\\
2. Always output step by step, and at the end of each step, describe what screen status you are expected to get.\\
3. Try to gather as much information/procedural knowledge as possible from the video.\\
4. If the video contains the entire task, describe the entire task.\\
5. If the video really does not contain anything meaningful, output "No information available".
\\

\#\#\# Task description:

\{task\}
\\

Initial screenshot:

$<$image of the initial screenshot$>$
\\

\#\#\# Output format

\{

~~~~"description": "The detailed knowledge you learned from the video about the given task."

\}
\end{promptbox}

\begin{promptbox}[label={real_scripts}]{Action Prediction with Video Guidance Prompt}
\{tool\_definitions\}\\

\{output\_format\_definitions\}\\

\#\#\# Task Instruction\\
\{task\}\\

\#\#\# The following describes the knowledge that supports this task from a tutorial video:\\
\{guidance\}\\
You can use it as the guide to perform the task.\\

\#\#\# History action sequence:\\
\{history\}
\end{promptbox}

\subsection{Video-based Verification Prompt}

\begin{promptbox}[label={real_scripts}]{Video-based Verification Prompt}
You are an expert GUI agent reviewer. You will watch the following video of an agent performing a task while describing its action every step.
\\

The task is:\\
\{task\}
\\

You need to evaluate the completion of this task, with the following score schemes:\\
0-1 | No meaningful progress; wrong app, crash, infinite repetition, or immediate failure.\\
2-3 | Correct environment but core action not executed; loops or early abandonment.\\
4-5 | Main action attempted; deliverable exists but is flawed or missing key constraints.\\
6-7 | Task complete with minor errors (e.g., slight formatting, naming) or inefficient path.\\
8-9 | Fully correct final state, all constraints satisfied, clean and robust execution.\\

You should focus on the following aspects in your evaluation:\\
- Final frames: Is the exact deliverable visible and correct (saved file, sent email, correct UI state)?\\
- Constraints: Were ALL specific requirements from the instruction fulfilled?\\
- Coherence: Penalize unresolved dialogs, unsaved files, and infinite loops.\\

Carefully analyze the video, then directly provide your final scores in the following format:\\
\{

~~~~"score": "your predicted score"

\}
\end{promptbox}

\subsection{Video Search and Relevance Scoring Prompt}

The prompt used to generate query and the coarse filtering of videos based on simple text-based queries. The description required is just downloaded description made by the content creators themselves from YouTube.

\begin{promptbox}[label={real_scripts}]{Video Guidance Extraction Prompt}
You are given the following computer use task instruction:\\
\{task\}\\

Generate 10 search queries to search tutorial videos online.\\

Output format:\\
\{

~~~~"queries": [

~~~~~~~~"query 1",

~~~~~~~~"query 2"

~~~~]

\}
\end{promptbox}

\begin{promptbox}[label={real_scripts}]{Video Relevance Filtering Prompt}
You are given the following computer use task instruction:\\
\{task\}\\

You are given the following video title and description:\\

Title:\\
\{title\}\\

Description:\\
\{description\}\\

Determine the relevance of the video in a scale of 0 to 5, where 0 means completely irrelevant, and 5 means exact task.
\end{promptbox}




\section{Example Guidance Output}
\label{sec:guidance}

\begin{figure*}[t]
    \centering
    \includegraphics[width=0.7\linewidth]{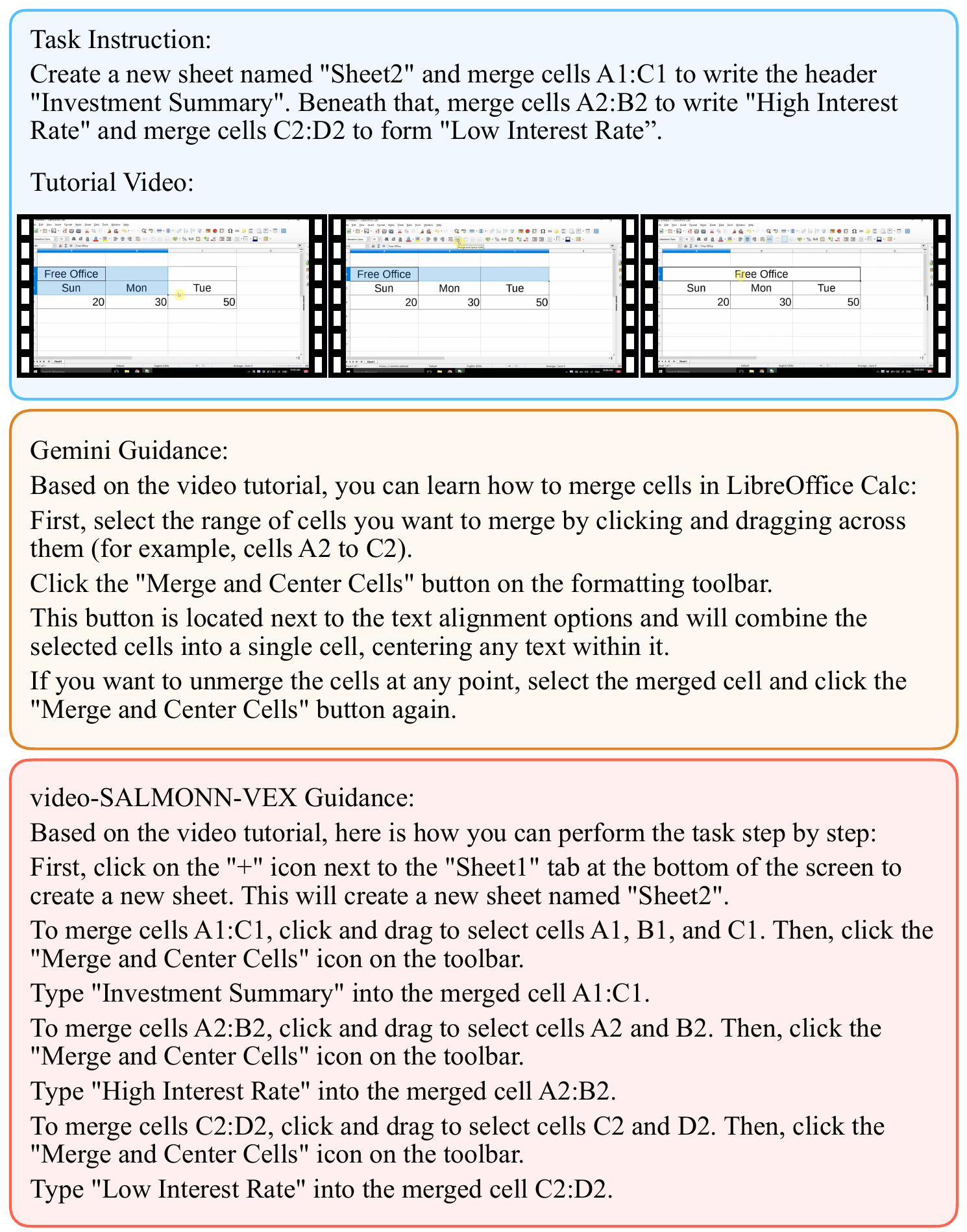}
    \caption{Example guidance generated from Gemini-3.5-flash and video-SALMONN-VEX for one OSWorld task. The figure shows a screenshot of the video, and the video's audio is also part of the input.}
    \label{fig:eg1}
\end{figure*}

\begin{figure*}[t]
    \centering
    \includegraphics[width=0.7\linewidth]{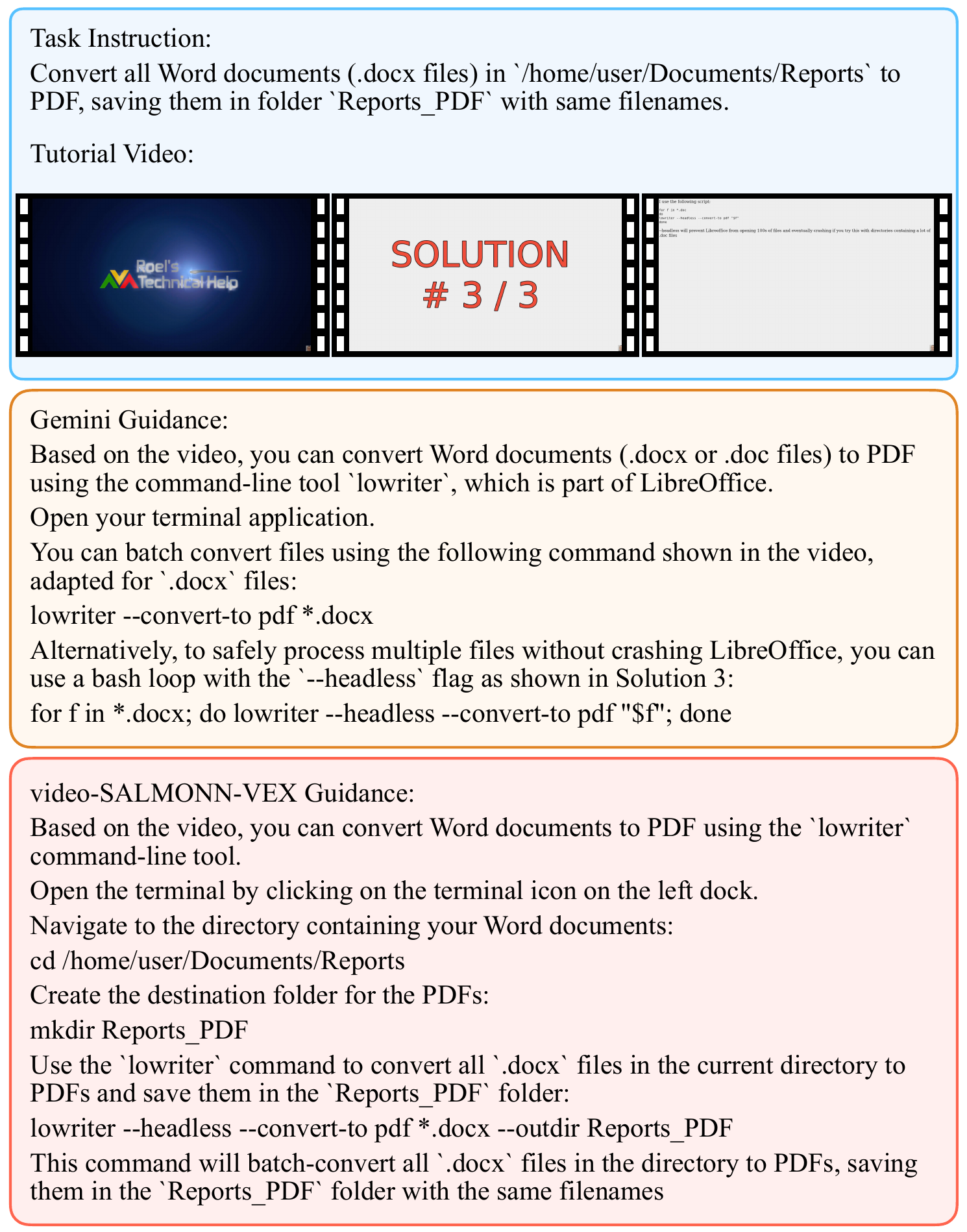}
    \caption{Example guidance generated from Gemini-3.5-flash and video-SALMONN-VEX for one OSWorld task. The figure shows a screenshot of the video, and the video's audio is also part of the input.}
    \label{fig:eg2}
\end{figure*}

Two examples of output guidance text from Gemini-3.5-flash and video-SALMONN-VEX are shown in Fig. \ref{fig:eg1} and \ref{fig:eg2}. The knowledge output from Gemini-3.5-flash are more faithful towards the video itself without paying much attention to the actual task setting. On the contrary, the output from video-SALMONN-VEX is more adapted to the specific task that we are performing. Both are effective with video-SALMONN-VEX guidance slightly outperforming Gemini-3.5-flash ones by 0.6\% using a Gemini-3.5-flash-based agent as shown in Table \ref{tab:osworldgemini}.

